\documentclass[10pt,twocolumn,letterpaper]{article}

\usepackage{wacv}              

\usepackage{graphicx}
\usepackage{amsmath}
\usepackage{amssymb}
\usepackage{booktabs}
\usepackage{csquotes}
\usepackage{marvosym}
\usepackage{tabularx}
\usepackage{color}
\usepackage{colortbl}
\usepackage{xcolor}
\usepackage{multirow}
\usepackage{multicol}
\usepackage{wrapfig}
\usepackage{algorithm}
\usepackage{algorithmic}

\usepackage[pagebackref,breaklinks,colorlinks]{hyperref}

\usepackage[capitalize]{cleveref}
\crefname{section}{Sec.}{Secs.}
\Crefname{section}{Section}{Sections}
\Crefname{table}{Table}{Tables}
\crefname{table}{Tab.}{Tabs.}

\def\wacvPaperID{1922} 
\def\confName{WACV}
\def\confYear{2025}

\begin{document}


\title{MagicStick: Controllable Video Editing via Control Handle Transformations}
\author{Yue Ma$^{1}$, \quad Xiaodong Cun$^{2\textrm{\Letter}}$,\quad  Sen Liang$^{3}$, \quad Jinbo Xing$^{5}$ \quad
  Yingqing He$^{1}$, \quad  Chenyang Qi$^{1}$, \\
  \quad  Siran Chen$^{4}$, \quad  Qifeng Chen$^{1\textrm{\Letter}}$ \\
  $^{1}$HKUST \quad  $^{2}$Great Bay University \quad 
  $^{3}$USTC \quad $^{4}$SIAT@MMLab \quad $^{5}$CUHK \\
\url{https://magic-stick-edit.github.io/}
}

\renewcommand{\thefootnote}{\fnsymbol{footnote}}


\twocolumn[{
\maketitle
\begin{center}
    \captionsetup{type=figure}
    \includegraphics[width=0.95\linewidth]{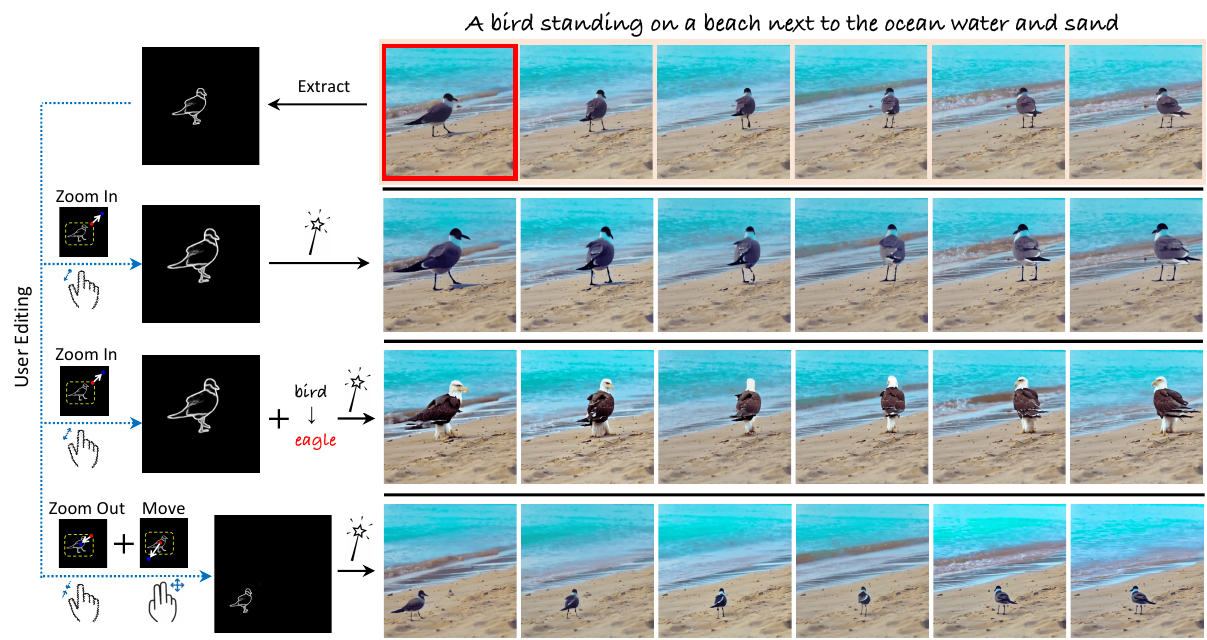}
    \captionof{figure}{\textcolor{black}{\textit{\textbf{ Controllable video editing via modifying control handle.}}} We present a unified framework to modify video properties~(\eg, shape, size, location, motion) leveraging the manual keyframe transformations on the extracted internal control signals.}
    \label{fig:teaser} 
\end{center}
}]

\newcommand\blfootnote[1]{%
  \begingroup
  \renewcommand\thefootnote{}\footnote{#1}%
  \addtocounter{footnote}{-1}%
  \endgroup
}

\footnotetext[2]{Equal contribution.} \footnotetext[0]{${\textrm{\Letter}}$ Corresponding author.}

\begin{abstract}
Text-based video editing has recently attracted considerable interest in changing the style or replacing the objects with a similar structure. 
Beyond this, we demonstrate that properties such as shape, size, location, motion, \textit{etc.}, can also be edited in videos.
Our key insight is that the keyframe's transformations of the specific internal feature~(\textit{e.g.}, edge maps of objects or human pose), can easily propagate to other frames to provide generation guidance.
We thus propose \texttt{MagicStick}, a controllable video editing method that edits the video properties by utilizing the transformation on the extracted internal control signals. 
In detail, to keep the appearance, we inflate both the pre-trained image diffusion model and ControlNet to the temporal dimension and train low-rank adaptions~(LoRA) layers to fit the specific scenes. 
Then, in editing, we perform an inversion and editing framework. Differently, finetuned ControlNet is introduced in both inversion and generation for attention guidance with the proposed attention remix between the spatial attention maps of inversion and editing. 
Yet succinct, our method is the ﬁrst method to show the ability of video property editing from the pre-trained text-to-image model. 
We present experiments on numerous examples within our uniﬁed framework. We also compare with shape-aware text-based editing and handcrafted motion video generation, demonstrating our superior temporal consistency and editing capability than previous works. 
\end{abstract}    
\section{Introduction}
Due to the remarkable progress of text-to-image (T2I) generation~\cite{he2022latent,ramesh2022hierarchical,ho2022imagen}, video editing has recently achieved significant progress leveraging the generation prior of text-to-image models~\cite{rombach2022high}.
Previous works have studied various editing effects, such as visual style transfer~\cite{geyer2023tokenflow, yang2023rerender, qi2023fatezero, bar2022text2live} or modification of the character and object to a similar one~\cite{chai2023stablevideo, lee2023shape, he2023animate}. 
However, many straightforward edits in videos remain out of reach currently. 

It appears feasible to edit images following the mentioned requirements as suggested by Epstein et al.~\cite{epstein2023selfguidance}, however, it proves challenging when applied to videos.
Firstly, the editing needs to be applied to all frames individually due to the distinction among video frames. Secondly, the temporal consistency will be deteriorated if we apply frame-wise editing to the video directly. Finally, taking resizing as an example, if we directly segment the object and then edit, the generated video still relies on the abilities of additional inpainting~\cite{avrahami2022blended,avrahami2023blended, xu2020e2i, yan2022texture} and harmonization~\cite{cun2020improving}. To handle these problems, our key insight is inspired by ControlNet~\cite{zhang2023adding}, where the structure can be generated following additional control. Editing the control signal is relatively easier and clearer than directly editing the appearance. 

\begin{figure}
    \centering
    \includegraphics[width=\linewidth]{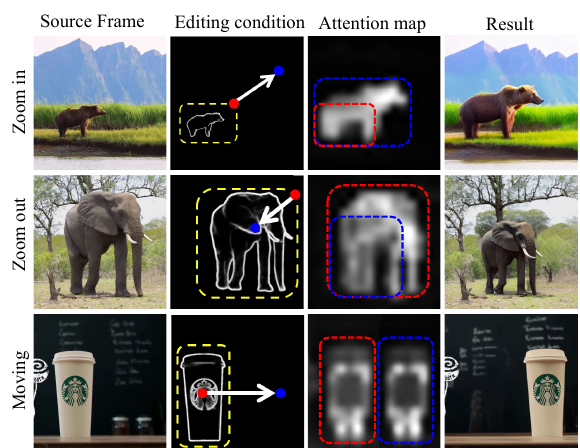}
    \vspace{-1em}
    \caption{\textit{\textbf{Motivation of the proposed method. }}
    We discover that the attention map can be edited through conditional editing. Therefore, we apply conditional injection to the inversion and denoising stages to provide better information guidance.
    }
    \label{fig:Motivation}
\end{figure}

As shown in the Fig.~\ref{fig:Motivation}, the results demonstrate that conditional injection plays a guiding role in the model's attention module.
To this end, we propose a universal video editing framework \texttt{MagicStick}. In addition to common video editing tasks (\eg, editing appearance), MagicStick is also capable of geometry editing, \eg, editing the object shape and position while maintaining its ID. To support geometry editing, we propose a two-stage `preserve-and-edit' strategy.
Specifically, to preserve the overall appearance, we first train a controllable video generation network following~\cite{ma2023follow, wu2023tune} based on the pre-trained Stable Diffusion~\cite{rombach2022high} and ControlNet~\cite{zhang2023adding} to gain the ability to generate videos with the specific appearance of the given video. After that, we involve the edited/transformed control signals in both video inversion~\cite{mokady2023null} and generation processes. 
Finally, we propose a novel attention remixing module, which performs editing by guiding the remixing of attention using fine-tuned ControlNet signals in both the inversion and generation processes. This dual-stage signal guidance significantly enhances the controllability of the editing process, enabling us to successfully modify specific aspects of a video, such as shape, size, and localization, as demonstrated in Fig.~\ref{fig:teaser}. Our method only requires adjusting a single video and does not need additional datasets for training. By editing the control signals of the video, we can achieve modifications to properties such as shape, size, location, and motion without the need for large-scale data training. This makes our method more efficient and widely applicable.
Besides the mentioned attributes, we can also alter the motion in the input human video, which is also absent in previous T2I-based video editing approaches.
To justify the effectiveness of our novel editing framework, we conduct extensive quantitative and qualitative experiments on various videos, demonstrating the superiority of our approach. 

Our contributions are summarized as follows: 

\begin{itemize}
    \item We demonstrate the ability of a series of important video editing aspects~(including shape, size, location, and human motion) for the first time by introducing a novel unified controllable video editing framework using pre-trained T2I models.
    \item We propose an effective attention remix module utilizing the attention from control signals to faithfully retain the edit-unrelated information from the source video. 
    \item The experiments show the advantages of the proposed method over the previous similar topics, \eg, shape-aware video editing and handcrafted motion controls.  
\end{itemize}

\section{Related Work}
\label{related_work}

\noindent\textbf{Video Editing.}
Editing natural videos is a vital task that has drawn current researchers' attention in computer vision. 
Before the advent of diffusion models, many GAN-based approaches~\cite{goodfellow2020generative, he2023gaia, fang2024real, ma2024followyourpose, ma2024followyourclick, ma2022visual, ma2024followyouremoji, chen2024follow, wang2024cove, gal2022stylegan, park2019semantic, patashnik2021styleclip} have achieved good performance. 
The emergency of diffusion models~\cite{song2020denoising} delivers higher quality and more diverse editing results. 
Text2live~\cite{bar2022text2live} and StableVideo~\cite{chai2023stablevideo} present layer-atlas-based methods and edit the video on a flattened texture map.
FateZero~\cite{qi2023fatezero} and Video-p2p~\cite{liu2023video}, guided by original and target text prompts, perform semantic editing by blending cross-attention activation maps in the denoising U-Net.   
There are also some approaches~\cite{chen2025livephoto, esser2023structure,molad2023dreamix, feng2024dit4edit, liu2024magicquill, guo2024refir} edit the appearance of videos by powerful yet private video diffusion models.
However, Most of these methods mainly focus on editing the texture rather than the shape, where the latter is more challenging.
They show obvious artifacts even with the optimization of generative priors. In contrast, our framework can achieve the editing of complex properties, including shape, size, and location of objects, while maintaining both appearance and temporal consistency.

\begin{figure*}
    \centering
    \includegraphics[width=1.0\linewidth]{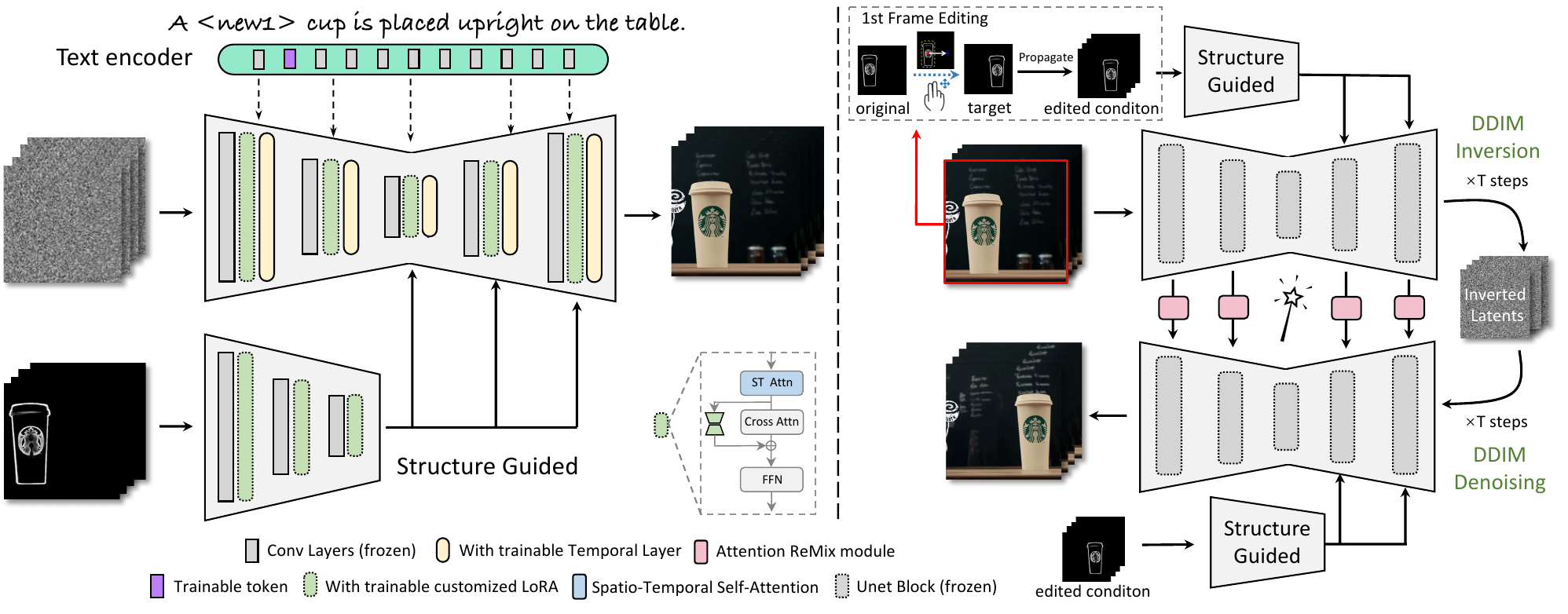}
    \vspace{-1em}
    \caption{\textbf{\textit{Overview of \texttt{MagicStick}.}} We propose a controllable video editing method that edits the video properties by utilizing the transformation on the extracted internal control signals. To achieve this, on the \textbf{left}, we tune the given video based on the extracted control signals and the text prompt for video customization. On the \textbf{right}, we first edit the key frame manually and propagate it to other frames as guidance. Then, relying on the proposed attention remix module, we can achieve controllable video editing using structure guidance during inversion and denoising.
    }
    \label{fig:framework}
    \vspace{-10pt}
\end{figure*}

\noindent\textbf{Image and Video Generation.}
Text-to-image generation is a popular topic with extensive research in recent years. 
Many approaches have been developed based on transformer architectures~\cite{ramesh2021zero, yu2022scaling, yu2021vector, ding2022cogview2, yan2021videogpt, hong2022cogvideo, ho2022imagen, he2023weaklysupervised, he2023camouflaged} to achieve textual control for generated content.
However, it operates attention during generation and struggles to maintain consistency with input. Moreover, it is designed for image generation and cannot be directly applied to video editing. 
To address a similar problem in video generation, various works~\cite{blattmann2023align, ma2022visual, liu2022towards, he2022latent, luo2023decomposed, ma2023follow, wang2024taming, zhang2023controlvideo, gao2024inducing} try to extend image LDM~\cite{rombach2022high} to video domain.
These methods then generate continuous content by incorporating additional temporal layers. Tune-A-Video~\cite{wu2023tune} proposes a method specializing in one-shot text-to-video generation tasks. The model has the ability to generate video with similar motion to the source video. However, how to edit real-world video content using this model is still under-exposed. 
Inspired by the image controllable generation methods~\cite{zhang2023adding, yang2023eliminating, yu2022pcfn, peng2021unified}, our method supports editing real-world video properties by leveraging pre-trained text-to-image models.

\section{Method}
\label{Method}

We aim to edit the property changes~(\eg, shape, size, location, and motion) in a video through transformations on one specific control signal as shown in Fig.~\ref{fig:framework}. Below, we start with a brief introduction to latent diffusion model (LDM) and inversion in Sec.~\ref{sec:preliminary}. Then, we introduce our video customization method in Sec.~\ref{sec:customization} to keep the appearance. Finally, we present the details of control signal transformation (Sec.~\ref{sec:condition_editing}) and editing during inference~(Sec.~\ref{sec:controllable Video Editing}).

\subsection{Preliminary}
\label{sec:preliminary}

\noindent\textbf{Latent Diffusion Models (LDMs).}
Derived from diffusion Models, LDMs~\cite{rombach2022high} reformulate the diffusion and denoising procedures within a latent space. Firstly, an encoder $\mathcal{E}$ compresses a pixel space image $x$ to a low-resolution latent $z=\mathcal{E}(x)$
, which can be reconstructed from latent feature to image $ \mathcal{D}(z) \approx x $ by decoder $\mathcal{D}$.
Then, a U-Net~\cite{ronneberger2015u} $\varepsilon_\theta$ 
with self-attention~\cite{vaswani2017attention} and cross-attention is trained to estimate the added noise using the following objective:
\vspace{-0.5em}
\begin{equation}
\label{eq:mse}
\min _\theta E_{z_0, \varepsilon \sim N(0, I), t \sim \text { Uniform }(1, T)}\left\|\varepsilon-\varepsilon_\theta\left(z_t, t, p \right)\right\|_2^2,
\end{equation}
where $p$ is the embedding of the text prompt and $z_t$ is a noisy sample of $z_0$ at timestep $t$. After training, we can generate clean image latents $z_0$ from random Gaussian noises $z_T$ and text embedding $p$ through step-by-step denoising and then decode the latents into pixel space by $\mathcal{D}$.


\noindent\textbf{DDIM Inversion.} During inference, we can use deterministic DDIM sampling to transform a random noise $z_T$ to a clean latent $z_0$ across a sequence of timesteps from $T$ to $1$:
\begin{equation}
\label{eq: denoise}
    z_{t-1} = \sqrt{\alpha_{t-1}} \; \frac{z_t - \sqrt{1-\alpha_t}{\varepsilon_\theta}}{\sqrt{\alpha_t}}+ \sqrt{1-\alpha_{t-1}}{\varepsilon_\theta},
\end{equation}
where $\alpha_{t}$ is the parameter for noise scheduling~\cite{ho2020denoising, song2020denoising}.
Thus, DDIM Inversion~\cite{dhariwal2021diffusion} is proposed to inverse the above progress from a clean latent space $z_0$ to a noisy latent space $z_T$ by adding noises:
\begin{equation}
\label{eq: add noise}
    \hat{z}_{t} = \sqrt{\alpha_{t}} \; \frac{\hat{z}_{t-1} - \sqrt{1-\alpha_{t-1}}{\varepsilon_\theta}}{\sqrt{\alpha_{t-1}}} + \sqrt{1-\alpha_{t}}{\varepsilon_\theta},
\end{equation}
In this context, $z_0$ can be reconstructed by inverted latent $\hat{z}_T$ using DDIM. 
By using DDIM inversion, we can edit real-world images and videos in a relatively deterministic way.

\subsection{Controllable Video Customization}  
\label{sec:customization}


Since the generation process of the generative model is too stochastic to keep the appearance, for our tasks, we first tune the network to satisfy our requirements, which is similar to model customization in text-to-image generation~\cite{ruiz2023dreambooth}.
Differently, we also involve the structure-guided model to add additional correspondence between the control signal and output, which is a trainable ControlNet~\cite{zhang2023adding} with additional designs of temporal attention layer~\cite{wu2023tune}, trainable LoRA~\cite{hu2021lora} and token embedding for better video customization fine-tuning. We follow the original MSE loss (Eq.~\ref{eq:mse}) for customization training.


In detail, as shown in Fig.~\ref{fig:framework}, the low-rank matrices~\cite{hu2021lora} are injected into pre-trained linear layers within the cross-attention modules of the denoising UNet. LoRA employs a low-rank learnable factorization technique to update the attention weight matrix $W_{q}$, $W_{k}$, $W_{v}$:
\begin{equation}
\begin{split}
W_{i}=W_{i_{0}}+\Delta W_{i_{0}}=W_{i_{0}}+B_{i}A_{i},
\end{split}
\end{equation}
where $i= q, k, v$ denotes the different part in cross-attention.
$W_{i_{0}} \in R^{d \times k}$ represents the original weights of the pre-trained T2I model~\cite{rombach2022high}, $B \in R^{d \times r}$ and $A \in R^{r \times k}$ represent the low-rank factors, where $r$ is much smaller than original dimensions $d$ and $k$. 
Remarkably, this operation does not affect the ability of the pre-trained T2I model~\cite{rombach2022high} to generate and compose concepts because LoRA introduces minimal, low-rank updates that preserve the core knowledge and structure of the model. We also train a token embedding to get better customization following \cite{gal2022image}.
In addition, we adopt inflated ControlNet~\cite{zhang2023adding} inspired by Tune-A-Video~\cite{wu2023tune} to include temporal information among the control signal sequences. 
The original control signal (\eg, sketch or edge, depending on the editing needs in the following sections) of the object is encoded by a structure-guided module, where we convert the original self-attention module in text-to-image diffusion U-Net backbone to spatio-temporal self-attention.

\subsection{Control Signal Transformation} 
\label{sec:condition_editing}
Since naive object editing in a video may deteriorate the surrounding background, we opt to edit the control signal of the first frame to ensure more consistent editing results. Then, the edited control signals are incorporated into U-Net after being encoded by ControlNet-based~\cite{zhang2023adding} structure guided module.
As shown in Fig.~\ref{fig:framework} (right), our control signal editing involves three steps. 

\noindent\textbf{1) Extraction.} We first segment the interest objects using a text-guided Segment-and-Track-Anything.~\cite{cheng2023segment}. Then the annotator~(\eg, pose detector, HED-detector, and depth estimator) is utilized to extract corresponding control signals from the clean segmented object sequences.

\noindent\textbf{2) Transformation.} After getting the intermediate representations of the specific frame, 
the user can perform single or multiple geometric editing operations on objects in the first signal frame, such as resizing and moving. 

\noindent\textbf{3) Propagation.} 
We compute the transformation parameters in the first edited signal frame. In detail, we detect the bounding box of the objects in the first original control signal frame and edited one. 
The movement of the bounding box's center point is calculated as the position parameter, and the changes in length and width are used as the resize parameters. 
Then, we apply these transformations across all frames, ensuring consistent guidance throughout the video.

\subsection{Controllable Video Editing} 
\label{sec:controllable Video Editing}
With guidance and the appearance customization, 
we can finally edit the video in our framework. Specifically, we conduct video inversion to obtain the intermediate features and then inject these features during the denoising stage, as shown in Fig.~\ref{fig:framework} (right). Unlike the previous unconditional video inversion and editing pipeline~\cite{qi2023fatezero}, our method guides both processes with the control signals and then performs editing via the proposed attention remix modules.

\begin{figure}
    \centering
    \includegraphics[width=\linewidth]{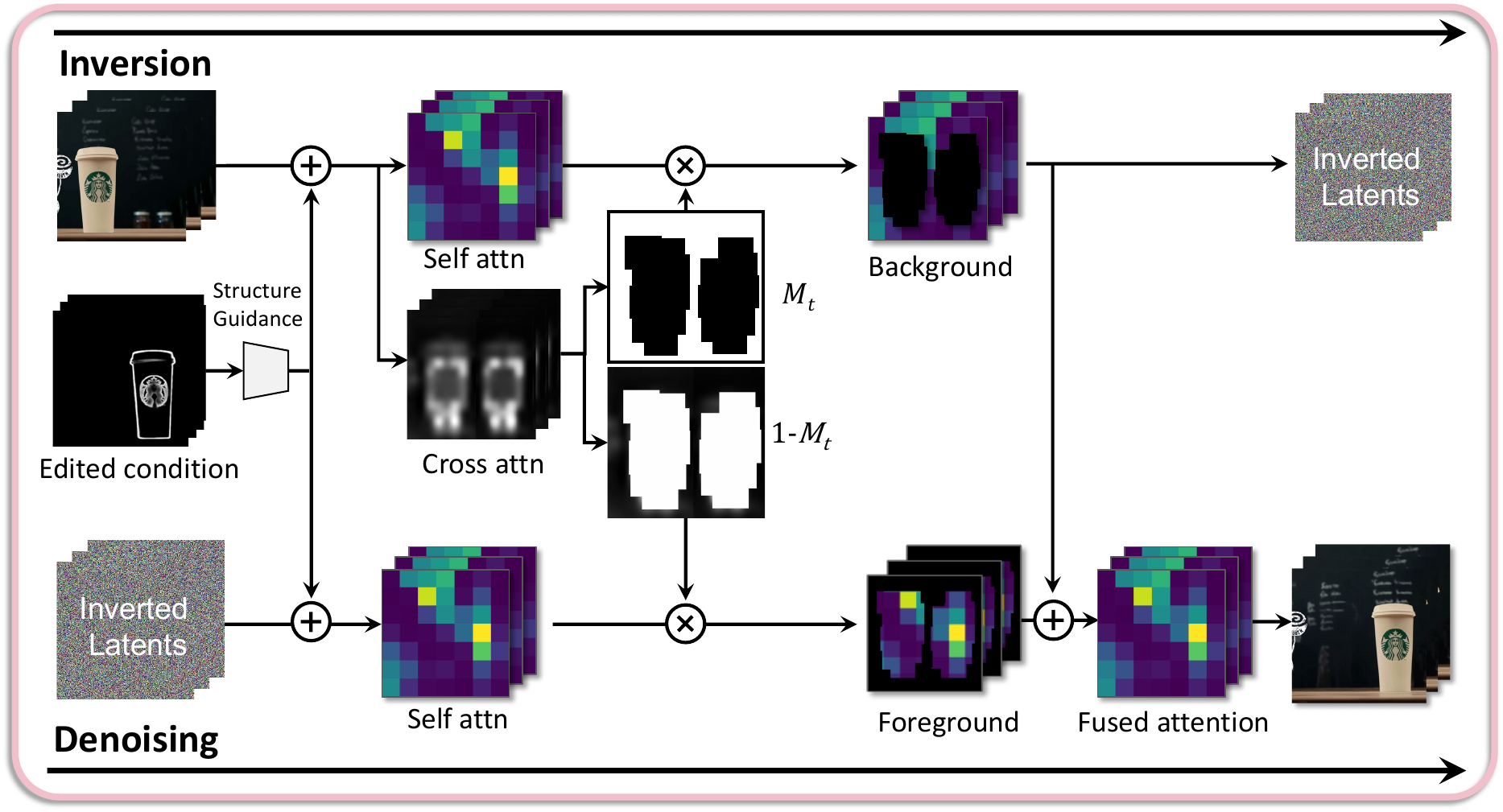}
    \vspace{-1em}
    \caption{\textit{\textbf{Attention ReMix module.}}
    Taking the moving editing as an example, we utilize edited conditions in both inversion and denoising. During inversion, the edited condition $\mathcal{C}^{edit}$ is employed to edit self-attention and cross attention from input video and get binary mask $M_{t}$. While for denoising, $\mathcal{C}^{edit}$ is leveraged to guide the appearance generation in the target region. 
     }
    \label{fig:attention_remix}
\end{figure}

\noindent\textbf{Controllable Inversion and Generation.} 
Most existing works~\cite{mokady2023null, brooks2023instructpix2pix, ju2023direct} conduct inversion followed by editing to achieve appearance editing. While in our task, we aim to achieve more complex property editing by incorporating the control signals during the inversion and generation. 
Specifically,
the same edited signal frame in two stages is employed for different purposes.
During inversion, we use the structure-guided model to encode the edited signal frames and add them into the U-Net as residuals.
This operation can easily inject the shape and position information of the edited signals into the self-attention and cross-attention processes of inversion of the source video. 
As shown in Fig.~\ref{fig:attention_remix}, taking the moving editing operation as an example, we observe that both the original and target positions are activated by words \enquote{cup} after injection in cross-attention. During generation, the edited signal frame is reintroduced into the network to serve as guidance for regenerating specific highlighted regions. In both stages, the injection of the edited signal frame is indispensable to accomplish our editing tasks through their mutual coordination.

\begin{figure*}
  \centering
    \includegraphics[width=1\linewidth]{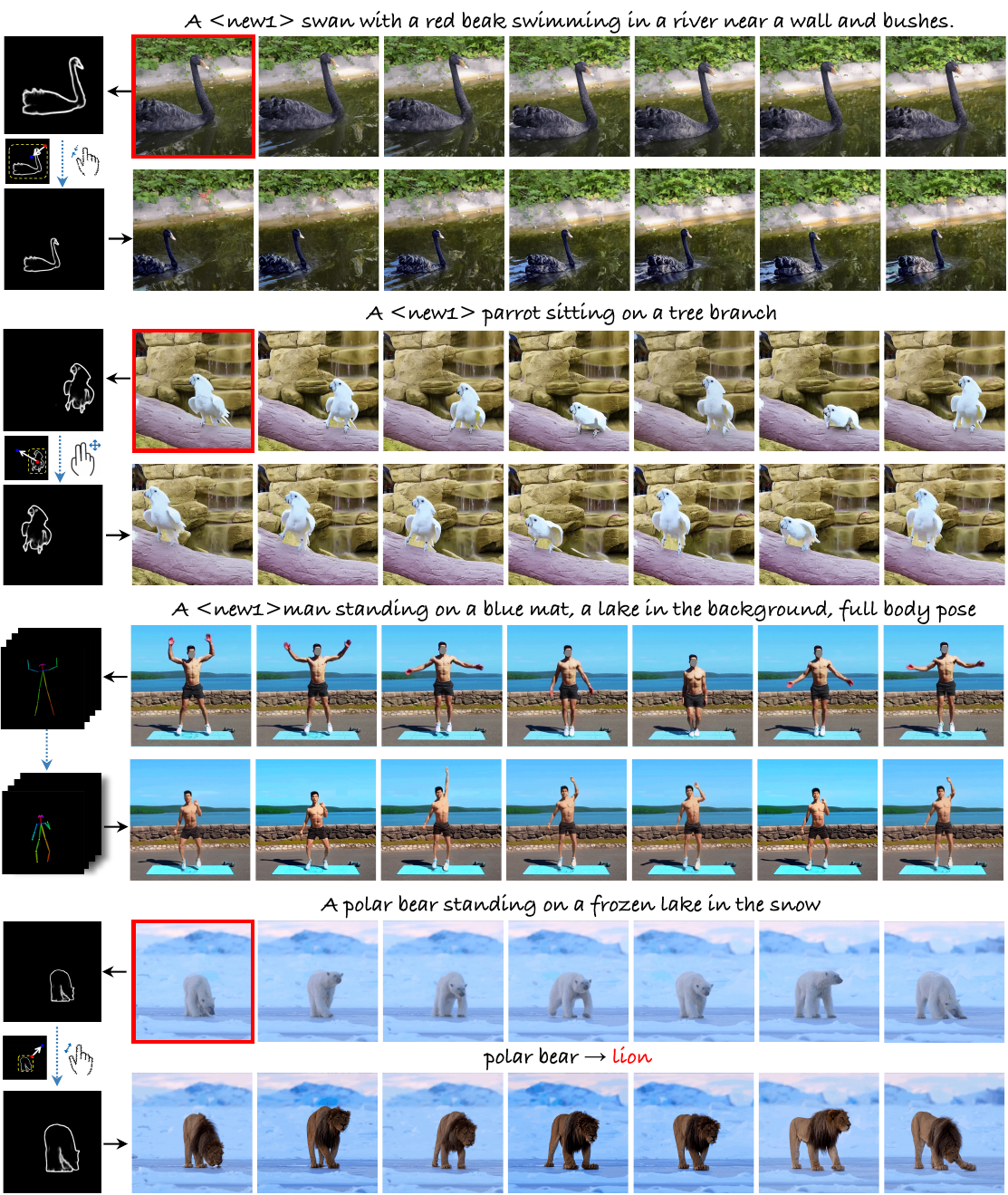}
  \caption{\textbf{Controllable video editing via modifying the control signals (\textit{e.g.,} sketch map and skeleton)}. Our framework can achieve consistent editing via the transformation of control signals while keeping its original appearance. We present the editing results of resizing~(small), moving~(transformation), human pose editing, and shape-aware text-based editing from top to bottom. We can further introduce text-based editing as the bottom sample to change both appearance and shape. Zoom in for the best view.
}
  \label{fig:application_single}
\end{figure*}

\begin{figure*}
    \centering
    \includegraphics[width=\linewidth]{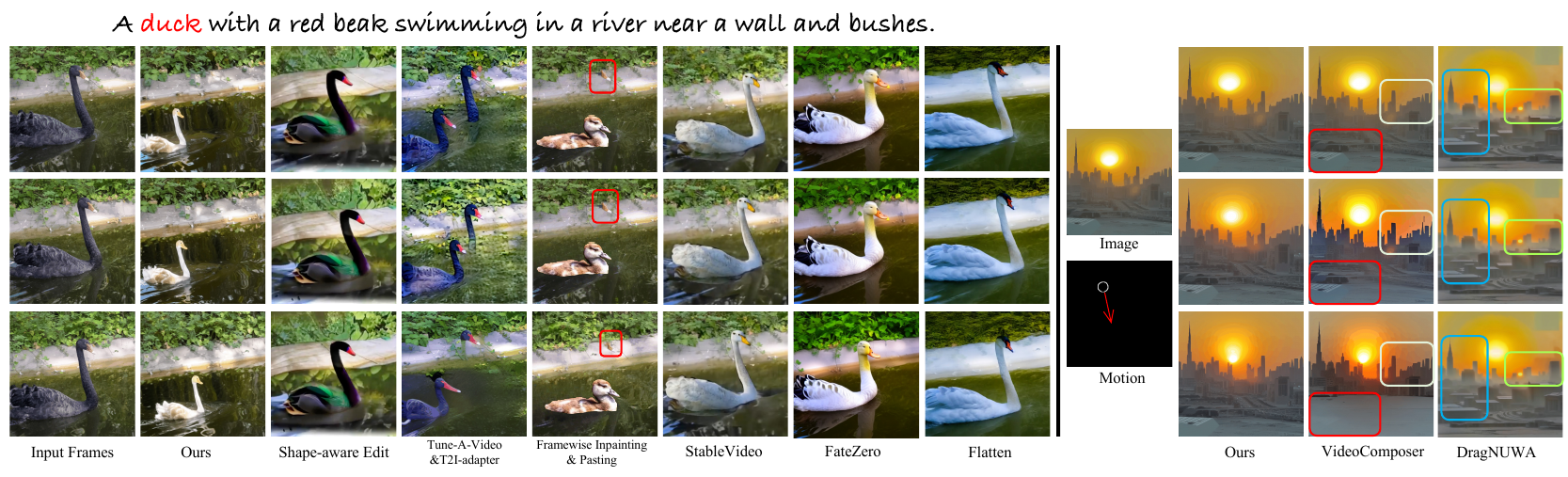}
    \vspace{-2em}
    \caption{\textit{\textbf{Qualitative comparison.}} \textbf{Left:} Shape-aware video editing. we resize the swan and replace word \enquote{\texttt{swan}} with \enquote{\texttt{duck}}. \textbf{Right:} Moving object, the first frame and motion conditions are provided to reconstruct the video. Our results exhibit the best temporal consistency, video ﬁdelity, and editing quality. Zoom in for the best view.
    }
    \label{fig:comparison}
    \vspace{-13pt}
    
\end{figure*}

\begin{figure*}
    \centering
    \includegraphics[width=\linewidth]{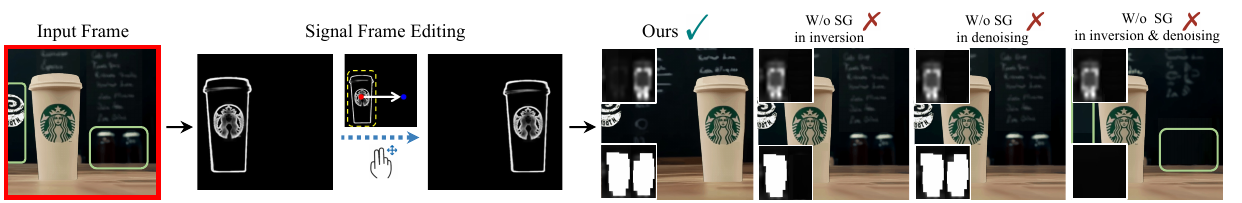}
    \caption{\textit{\textbf{Ablation about Attention ReMix module.}} SG indicates the structure guidance model.
    The sub-figures located in the upper-left corner of each editing result are the visualization of cross-attention activation maps during denoising, while the mask figures $1- M_{t}$ in the bottom-left corner are generated by the word activation of
    cross-attention in inversion. If we remove the structure-guided module in inversion or denoising from our model, it fails to complete the moving task (\enquote{$\checkmark$} represents the successful completion of the task, while \enquote{$\times$} indicates the failure).
    }
    \label{fig:ablation_attention_remix_module}
\end{figure*}

\noindent\textbf{Attention Remix Module.}
Adopting only the guidance of structure in inversion cannot achieve our novel editing tasks, as shown in the 3rd column of the editing results in Fig.\ref{fig:ablation_attention_remix_module}. 
To achieve our ultimate goal, we propose the Attention ReMix module to modify the attention in both inversion and denoising processes. As shown in Fig.~\ref{fig:attention_remix}, we store the intermediate self-attention maps $\{s_{t}^{\text{src}}\}_{t=1}^T$ and cross-attention maps $\{c_{t}^{\text{src}}\}_{t=1}^T$ at each timestep $t$ and the last noisy latent maps $z_T$ as:
\begin{equation}
z_T, \{c_{t}^{\text{src}}\}_{t=1}^T, \{s_{t}^{\text{src}}\}_{t=1}^T=\texttt{Inv}\left({z_0}, {\mathcal{C}}^{edit} \right),
\end{equation}
where $\texttt{Inv}$ donates the DDIM inversion pipeline and ${\mathcal{C}}^{edit}$ represents edited signal frame. 
As shown in Fig.~\ref{fig:attention_remix}, during the denoising stage, 
the activation areas of cross-attention by words \enquote{cup} provide significant assistance for binary mask $M_{t}$ generation, which is obtained by thresholding the cross-attention map $\{c_{t}^{\text{src}}\}_{t=1}^T$. Then we blend the $ \{s_{t}^{\text{src}}\}_{t=1}^T$ with $M_{t}$ and use the structure-guided module to generate the target object. Formally, this process can be represented as
\begin{equation}
\begin{aligned}
M_t &= \texttt{GetMask} \left( c_t^{\text{src}},  \tau \right),\\
s_{t}^{\text{final}} &= \texttt{G} \left( M_t \odot s_{t}^{\text{src}}, {\mathcal{C}}^{{edit}} \right) + (1 - M_t) \odot s_{t}^{\text{src}},
\end{aligned}
\end{equation}
where $\tau$ stands for the threshold value, $\texttt{GetMask}\left( \cdot \right)$ represents the operation to get mask using a specific word, and $\texttt{G} \left( \cdot  \right) $ denotes structure guidance.

\section{Experiments}
\label{Experiment}
\subsection{Implementation Details}
We implement our method based on the pre-trained Stable Diffusion (SD)~\cite{rombach2022high} and ControlNet~\cite{zhang2023adding} at 100 iterations. We sample 8 uniform frames at the resolution of $512\times512$ from the input video and ﬁne-tune the models for about 5 minutes on a single 24G NVIDIA RTX 3090Ti GPU. The time cost of training stage is cheap.

\subsection{Applications}
\noindent\textbf{Object size modification.} Using the pre-trained text-to-image diffusion model~\cite{rombach2022high}, our method supports object size modification by manually modifying the specific sketch in the first frame. 
As shown in the 1st and 2nd rows of Fig.~\ref{fig:application_single}, our method achieves consistent editing of the foreground \enquote{\texttt{swan}} by scaling down its sketch, while preserving the original background content.

\noindent\textbf{Object position editing.} One of the applications of our method is to edit the position of objects by adjusting the object's position in the first frame of the control signal.  
This task is challenging because the model needs to generate the object at the target area while inpainting the original area simultaneously. 
As demonstrated in the 3rd and 4th rows of Fig.~\ref{fig:application_single}, thanks to the proposed Attention ReMix module, our approach enables moving the \enquote{\texttt{parrot}} from the right of the branch to left by shifting the position of the parrot sketch in the 1st signal frame. The background and the object successfully remain consistent across different video frames.

\noindent\textbf{Human motion editing.} Our approach is also capable of editing human motion by replacing the source skeleton signals with the target sequences extracted from other videos. For instance, we can modify man's motion from \enquote{\texttt{jumping}} to \enquote{\texttt{raising hand}}, as shown in the 5th and 6th rows in Fig.~\ref{fig:application_single}. The result indicates that we can generate new content using target pose sequences while maintaining the human appearance in the source video (\eg, white shoes and black pants).


\noindent\textbf{Object appearance editing.}
Thanks to the rich knowledge learned in the per-trained T2I models~\cite{rombach2022high}, we can modify the object appearance through the editing of source text prompts, which is similar to existing mainstream video editing methods. Differently, we can further fulfill the property and appearance editing simultaneously. The 7th and 8th rows in Fig.~\ref{fig:application_single} showcase this powerful and intriguing capability.  We replace \enquote{\texttt{bear}} with \enquote{\texttt{lion}} simply by modifying the corresponding words and enlarging its sketch. 


\subsection{Comparisons}
\label{sec:Baseline Comparisons}
We compare our method with other video editing approach, including Shape-aware Edit~\cite{lee2023shape}, Tune-A-Video~\cite{wu2023tune}, StableVideo~\cite{chai2023stablevideo}, FateZero~\cite{qi2023fatezero} and Flatten~\cite{cong2023flatten}. 
We report the qualitative and quantitative results, respectively. 
The specific editing task for comparison is simultaneously editing the shape and appearance. 

\noindent\textbf{Qualitative results.} We present various results from different open-source approaches.
As shown in the 3rd and 5th columns in Fig~\ref{fig:comparison}.
The first is Shape-aware editing~\cite{lee2023shape}, its appearance is still limited by the optimization, causing blur and unnatural results. And the second one is to segment the objects and direct inpainting~\cite{yu2023inpaint} and paste directly. However, we find this naive methods struggles with the video harmonization and temporal artifacts caused by inpainting and segmentation. Another relevant method is using Tune-A-Video~\cite{wu2023tune} with T2I-Adapter~\cite{mou2023t2i}. It accomplishes local editing after overfitting but still exhibits temporal inconsistency.
We also compare the method with~\cite{chai2023stablevideo, qi2023fatezero,cong2023flatten}, they are difficult to generate realistic video while modifying the shape.
Differently, our proposed methods manage to resize the \enquote{\texttt{swan}},\enquote{\texttt{truck}} and edit its appearance to \enquote{\texttt{duck}}, \enquote{\texttt{train}} simultaneously, while maintaining the temporal coherence. 
On the other hand, we compare our method on conditional video generation from handcrafted motion signals and a single appearance. As shown in Fig.~\ref{fig:comparison}, VideoComposer~\cite{wang2023videocomposer} struggles to generate temporally consistent results, while DragNUWA~\cite{yin2023dragnuwa} fails to preserve the visual details of the input image.
In contrast, our method can produce video results with natural appearance and better temporal consistency.



\noindent\textbf{Quantitative results.} 
We also conduct quantitative comparisons with the following metrics: 
1) \textit{Tem-Con} (Temporal Consistency): Following the previous methods~\cite{esser2023structure, wang2023gen, xing2023make}, we evaluate temporal consistency of the generated video frames by calculating the cosine similarity between all pairs of consecutive frame embeddings of CLIP~\cite{radford2021learning} image encoder.
2) \textit{Fram-Acc} (Frame Accuracy): For object appearance editing, we measure the frame-wise editing accuracy based on the editing text prompt, which is the percentage of frames where the edited frames have a higher CLIP similarity to the target prompt than the source prompt.
3) \textit{Four user studies metrics}: Following FateZero~\cite{qi2023fatezero}, we assess our approach using four user studies metrics (\enquote{\textit{Edit}}, \enquote{\textit{Image}}, \enquote{\textit{Temp}} and \enquote{\textit{ID}}).  They are editing quality, overall frame-wise image ﬁdelity, temporal consistency of the video, and object appearance consistency, respectively. For a fair comparison, we ask 30 subjects to rank different methods. Each study displays four videos in random order and requests the evaluators to identify the one with superior quality. 
As demonstrated in Tab.~\ref{table:Quantitative_baseline}, the proposed method achieves the best Tem-Con and Fram-Acc against baselines.
As for the user studies, the average ranking of our method earns user preferences the best in three aspects and the comparable performance in ID metric.

\begin{table}[t]
\caption{Quantitative evaluation on video editing against baselines. The best results are marked in bold.}
\resizebox{\linewidth}{!}{
\setlength{\tabcolsep}{2.1pt}
\centering
\begin{tabular}{@{}l@{\hspace{2mm}}c@{\hspace{2mm}}*{3}{c@{\hspace{2mm}}}cc@{\hspace{2mm}}c@{}}
\toprule
Method & \multicolumn{2}{c}{CLIP Metrics$\uparrow$} & \multicolumn{4}{c}{User Study$\downarrow$}  \\
\cmidrule(l{1mm}r{1mm}){2-3} 
\cmidrule(l{1mm}r{1mm}){4-7} 
Inversion \& Editing & Tem-Con & Fram-Acc & Edit  & Image & Temp & ID\\
\midrule
Tune-A-Video~\cite{wu2023tune} \& T2I-adapter~\cite{mou2023t2i} & 0.891 & 0.851 & 2.87 & 2.99 & 2.67 & 3.18 \\
Shape-aware Edit~\cite{lee2023shape} & 0.722 & 0.618 & 3.56 & 3.69 & 3.74 & 3.53 \\
VideoComposer~\cite{wang2023videocomposer} & 0.914 & 0.846 & 3.38 & 2.74 & 2.94 & 4.31 \\
Single-frame Inpainting \& Pasting & 0.920 & 0.887 & 3.92 & 3.97 & 4.08 & \textbf{1.96} \\
StableVideo~\cite{chai2023stablevideo} & 0.922 & 0.902 & 2.74 & 3.97 & 3.87 & 2.95 \\
FateZero~\cite{qi2023fatezero} & 0.917 & 0.884 & 3.92 & 2.42 & 3.16 & 2.74 \\
Flatten~\cite{cong2023flatten} & 0.919 & 0.913 & 3.92 & 2.26 & 2.52 & 2.87 \\
\midrule
Ours & \textbf{0.928} & \textbf{0.919} & \textbf{1.27} & \textbf{1.62} & \textbf{1.58} & 2.04 \\
\bottomrule
\end{tabular}
}
\vspace{1pt}
\label{table:Quantitative_baseline}
\end{table}

\begin{figure}
    \centering
    \includegraphics[width=\columnwidth]{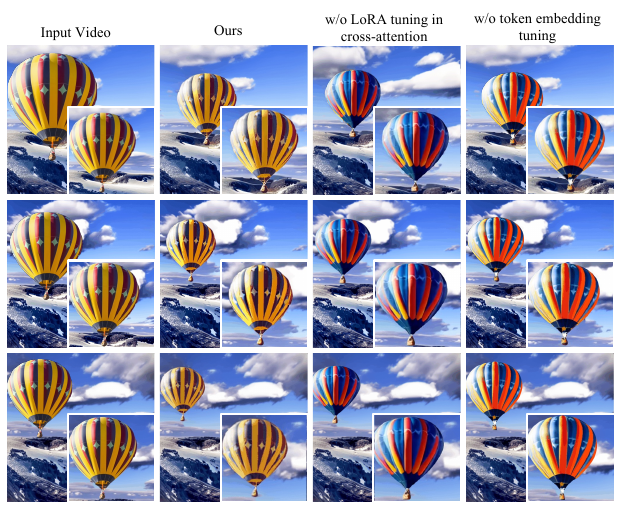}
    \caption{\textit{\textbf{Ablation on Video Customization.}}. Without LoRA tuning in cross-attention, the generated video can not preserve the object's content. Similarly, the appearance of the object~(\eg, texture and color) experiences a degradation in content consistency when token embedding tuning is absent.
    }
    \label{fig:ablation_customization_module}
\end{figure}

\subsection{Ablation Studies}

\noindent\textbf{Video Customization via LoRA and token embedding tuning.} 
We present the visual results of our variants, \ie, ours without LoRA in cross-attention or token embedding tuning in the 3rd and 4th column of Fig.~\ref{fig:ablation_customization_module}, respectively. The result in the 3rd column illustrates the significance of cross-attention LoRA for preservation of object appearance.
Additionally, we notice a deterioration in performance when token embedding remains untuned, which further underscores the significance of this process in content preservation. In contrast, our full method guarantees the preservation of texture and appearance consistency by meticulously fine-tuning both components.

\begin{table}[t]
\caption{Quantitative evaluation about ablation study. The best results are marked in bold. 
}
\resizebox{\linewidth}{!}{
\setlength{\tabcolsep}{2.1pt}
\centering
\begin{tabular}{@{}l@{\hspace{2mm}}c@{\hspace{2mm}}*{3}{c@{\hspace{2mm}}}cc@{\hspace{2mm}}c@{}}
\toprule
Method & \multicolumn{2}{c}{CLIP Metrics$\uparrow$} & \multicolumn{4}{c}{User Study$\downarrow$}  \\
\cmidrule(l{1mm}r{1mm}){2-3} 
\cmidrule(l{1mm}r{1mm}){4-7} 
Inversion \& Editing & Tem-Con & Fram-Acc & Edit  & Image & Temp & ID\\
\midrule
\small{w/o LoRA tuning in cross-attention}  & 0.891 & 0.884 & 3.27 & 3.34 & 2.91 & 3.97 \\
\small{w/o token embedding tuning }  & 0.908 & 0.915 & 2.65 & 3.09 & 2.63 & 3.82 \\
\small{w/o spatio-temporal self-attention} & 0.752 & 0.869 & 4.34 & 3.71 & 4.09 & 3.41 \\
\small{w/o temporal layer} & 0.884 & 0.893 & 3.52 & 3.12 & 3.92 & 2.40 \\
\midrule
Ours & \textbf{0.931} & \textbf{0.922} & \textbf{1.23} & \textbf{1.74} & \textbf{1.46} & \textbf{1.41} \\
\bottomrule
\end{tabular}
}
\label{table:Quantitative_baseline_ablation}
\end{table}

\noindent\textbf{Attention Remix Module.} We further investigate the effectiveness of the proposed Attention Remix module. The visual results of the ablation experiments are shown in Fig.~\ref{fig:ablation_attention_remix_module}, where we remove the structure-guided module to ablate its role during inversion and generation.
As presented in the 2rd column, removing the structure-guided module in inversion leads to the failure of moving editing. Since target area mask (bottom left corner) is missing, the self-attention in this region entirely derives from that of the background.
The 4th column demonstrates the results of an absence of structure guidance during generation.
This variant also fails to achieve the moving editing task even with the target area mask. 
This is because there is no guidance for the target area during the generation (upper-left corner of the 4th column). 
Finally, we also show the result produced by the variant by removing the attention remix module in both two stages. This ablated framework severely degrades to be a self-attention reconstruction. As shown in the 5th column, this variant fails to accomplish the moving task and maintain background consistency (green rectangles in the 5th column).
In contrast, when we equip the guidance both in inversion and generation, the \enquote{\texttt{cup}} can be shifted successfully, further emphasizing our module's significance. when we remove the Attention ReMix module, the position of cup is not changed, which evident the importance of proposed module. Note that the binary mask $M_{t}$, obtained by mapping between word and cross attention, is accurate since the utilization of powerful pretrained T2I model~\cite{rombach2022high}.

\section{Conclusion}
In this paper, we propose a new controllable video editing method \texttt{MagicStick} that performs temporally consistent video property editing such as shape, size, location, motion, or their combinations. To the best of our knowledge, our method is the first to demonstrate the capability of editing shape, size and localization of objects in videos using a pre-trained text-to-image model. To achieve this, we make an attempt to study and utilize the transformations on one control signal~(\eg, edge maps of objects or human pose) using customized ControlNet. A novel Attention ReMix module is further proposed for more complex video property editing. Our framework leverages pre-trained image diffusion models for video editing, which we believe will contribute to numerous potential video applications.


\paragraph{Acknowledgments.}
We thank Jiaxi Feng, Yabo Zhang for their helpful comments. 
This project was supported by the National Key R\&D Program of China under grant number 2022ZD0161501.

{\small
\bibliographystyle{ieee_fullname}
\bibliography{egbib}
}
\end{document}